\documentclass[conference]{IEEEtran}
\usepackage{ifpdf}

\usepackage{cite}

\ifCLASSINFOpdf
  \usepackage[pdftex]{graphicx}
\else
  \usepackage[dvips]{graphicx}
\fi
\usepackage{amsmath}
\usepackage{algorithm}
\usepackage{mathtools}
\usepackage{amssymb}

\usepackage{algorithmic}

\usepackage{array}

\ifCLASSOPTIONcompsoc
\usepackage[caption=false,font=normalsize,labelfont=sf,textfont=sf]{subfig}
\else
\usepackage[caption=false,font=footnotesize]{subfig}
\fi
\usepackage{fixltx2e}

\usepackage{stfloats}
\usepackage{url}

\usepackage{multicol}
\usepackage{multirow}
\usepackage{booktabs}

\begin{document}
%

\title{EntailE: Introducing Textual Entailment in Commonsense Knowledge Graph Completion}

\author{Ying Su, Tianqing Fang, Huiru Xiao, Weiqi Wang, \\
Yangqiu Song, Tong Zhang, Lei Chen \\
Hong Kong University of Science and Technology \\
ysuay@connect.ust.hk, tfangaa@cse.ust.hk, huiruxiao@ust.hk, wwangbw@cse.ust.hk, \\ yqsong@cse.ust.hk, tongzhang@ust.hk, leichen@cse.ust.hk}

\maketitle


\begin{abstract}
    Commonsense knowledge graph completion is a new challenge for commonsense knowledge graph construction and application.
    In contrast to factual knowledge graphs such as Freebase and YAGO, commonsense knowledge graphs (CSKGs; e.g., ConceptNet) utilize free-form text to represent named entities, short phrases, and events as their nodes.
    Such a loose structure results in large and sparse CSKGs,
    which makes the semantic understanding of these nodes more critical for learning rich commonsense knowledge graph embedding.
    While current methods leverage semantic similarities to increase the graph density, the semantic plausibility of the nodes and their relations are under-explored.
    Previous works adopt conceptual abstraction to improve the consistency of modeling (event) plausibility, but they are not scalable enough and still suffer from data sparsity.
    In this paper, we propose to adopt textual entailment to find implicit entailment relations between CSKG nodes, to effectively densify the subgraph connecting nodes within the same conceptual class, which indicates a similar level of plausibility.
    Each node in CSKG finds its top entailed nodes using a finetuned transformer over natural language inference (NLI) tasks, which sufficiently capture textual entailment signals.
    The entailment relation between these nodes are further utilized to: 1) build new connections between source triplets and entailed nodes to densify the sparse CSKGs; 2) enrich the generalization ability of node representations by comparing the node embeddings with a contrastive loss.
    Experiments on two standard CSKGs demonstrate that our proposed framework EntailE can improve the performance of CSKG completion tasks under both transductive and inductive settings.
\end{abstract}
\section{Introduction}

Understanding commonsense knowledge is critical for human-level artificial intelligence \cite{davis2015commonsense}. Commonsense knowledge graphs (CSKG) provide background knowledge for various downstream tasks such as question answering~\cite{sap-etal-2019-social}, dialogue responses~\cite{DBLP:conf/aaai/YoungCCZBH18}, and commonsense fact linking~\cite{gao2022comfact}. 
However, commonsense Knowledge Graphs (e.g., ConceptNet \cite{speer2017conceptnet} and ATOMIC \cite{sap2019atomic}) are highly incomplete due to the complexity of natural language and sparsity of knowledge and human annotations~\cite{malaviya2020commonsense}. Therefore it is an important task to predict missing links for such graphs. 


Commonsense knowledge graph completion poses significant challenges due to the sparse graph structures and complex node semantics in CSKGs~\cite{malaviya2020commonsense}. Though conventional knowledge graph completion methods, such as ConvTransE~\cite{shang2019end}, are often effective for dense factual knowledge graphs such as WordNet \cite{DBLP:journals/cacm/Miller95wordnet} and Freebase \cite{DBLP:conf/sigmod/BollackerEPST08}, they still struggle with completing CSKGs due to the aforementioned challenges~\cite{malaviya2020commonsense}.

\begin{figure}[t]
\centering
\includegraphics[scale=0.65,trim={0.5cm 0cm 0cm 0cm}]{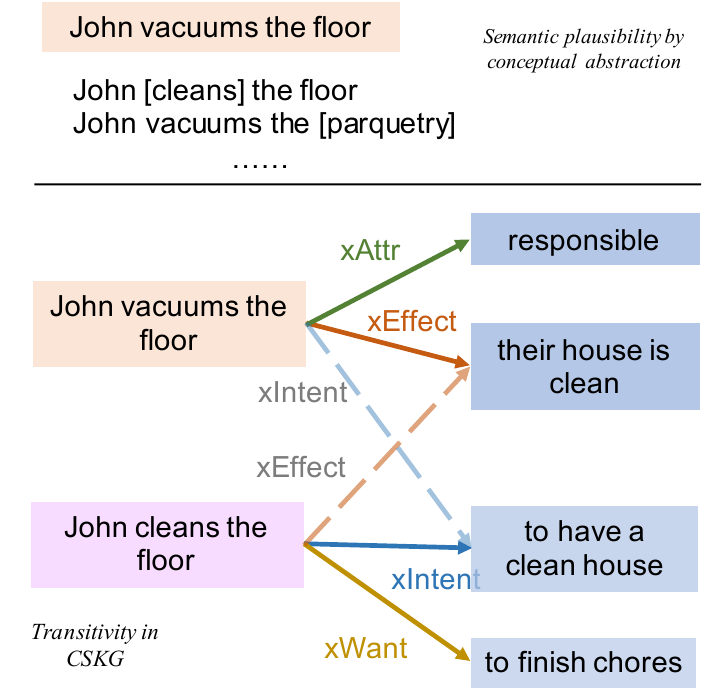}
\caption{
An example of densifying ATOMIC with nodes which are semantically related. Solid lines indicate the original edges and dashed lines are the densified edges using commonsense plausibility. 
}
\label{fig:intro}
\end{figure}

Prior research on CSKG completion has addressed the issue of sparsity by generating synthetic edges between similar nodes to densify the graph \cite{malaviya2020commonsense, wang2020inductive}. Equipped with graph neural networks \cite{kipfsemi, schlichtkrull2018modeling}, synthetic edges are only used for the interaction and updating of node representations without involving in the final triplet scoring directly. 
However, such embedding similarity does not bring in new knowledge, i.e., new triplets as defined in the original paradigm of the CSKG, which can also be useful in enhancing the representation learning. 

Semantic plausibility describes the \textit{selectional preference} of human ability in understanding natural language \cite{wilks1975preferential}, the preference of a predicate taking an argument to form a particular dependency pattern, e.g., the predicate ``vacuum'' and the argument ``the floor''. It is also a necessary component in commonsense related tasks \cite{pantel2007isp, zhang2019knowledge}. 
In CSKG completion, it's thus a natural way to densify the graph by mining semantic relations between CSKG node contexts based on their semantic plausibility among predicates and arguments, where nodes with a similar semantic plausibility level tends to have similar causes and effects.

Typical methods for modeling semantic plausibility of events under different lexical hierarchies are conducted from the perspective of conceptual abstraction \cite{2012Probase, yuenriching, porada2021modeling}. However, such methods require fine-grained extraction and matching of predicate-argument structure to provide commonalities between contexts. It is hard to directly apply the semantic plausibility for CSKG completion due to its various context form, including named entities, short phrases, and events.

Alternatively, to effectively identify node contexts under the same conceptual class, we resort to the idea of textual entailment, a key property to reflect how humans abstract the eventualities and describes whether one eventuality has more general meaning \cite{zacks2001event}. 
Besides this, textual entailment and transitivity alleviates the sparsity problem and benefits the entailment graph building on large-scale \cite{chen2022entailment}. Inspired by these, we can adopt textual entailment to model the semantic plausibility between CSKG node contexts and reformalize the transitivity for CSKG completion.\\

\noindent \textbf{Example 1:} Figure~\ref{fig:intro} illustrates an example of utilizing semantic plausibilty to densify ATOMIC. With conceptual abstraction, the plausibility is consistent between ``John vacuums the floor'' and ``John cleans the floor'' or ``John vacuums the parquery''. The conceptual abstraction is conducted by substituting ``over ``vaccum'' or ``floor'' with other concepts. By transiting the knowledge triplets from ``John vacuums the floor'' to ``John cleans the floor'' (also a context node in ATOMIC), the graph is densified. 

To facilitate the feasibility for extracting entailment relation over large scale graphs with various forms, embeddings for the node contexts are extracted with a finetuned transformer over NLI task \cite{maccartney2008modeling}. The pairwise entailment score is calculated by comparing the distances between node embeddings, satisfying the fact that plausibility is not strictly monotonic \cite{rudinger2020thinking, porada2021modeling}. The pairwise score indicates the extent of semantic plausibility between two CSKG nodes.
With the entailment relation, we manage to answer two questions: 
1) how to utilize the entailment relations to densify current sparse CSKGs?
2) how to enhance the node representation learning avoiding noises in the denfisying process?

For the first question, we define a novel transitivity between CSKG nodes and triplets: if $h \rightarrow h'$, then $(h,r,t) \rightarrow (h',r,t)$, $\rightarrow$ means ``entails''. The synthetic triplets $(h',r,t)$ are added to densify the graph, which transfer the annotation from $h$ to $h'$ to enhance the representation learning of both relation and context node. With the entailment relation between CSKG nodes, the CSKG triplets from the source node are transited to its entailed nodes, constructing new synthetic triplets to densify the CSKGs. In order to maintain the completeness of original graph, the synthetic triplets are not merged with the original triplets.


For the second question, since the synthetic triplets may introduce noises, we also create a dedicated entity contrast module for the purpose of leveraging entailment relations to facilitate entity embedding learning.
Entailment relations are naturally weak-supervised signals for grouping semantically related nodes while keeping away the unrelated ones. This benefits more to the isolated nodes with no training triplets but with entailment relation connecting the seen nodes in training set.

Our contributions are as follows:
\begin{itemize}
    \item We propose to adopt the semantic plausibility of node contexts in CSKG with textual entailment from finetuned transformer over NLI task, exploring the implicit correlation between the nodes;
    \item We propose to conduct CSKG completion with synthetic triplets constructed by transferring the triplets from source node to its entailed nodes. The synthetic triplets densify the original CSKGs and benefit representation learning, further improve the CSKG completion performance;
    \item To adopt the entailment relation for node representation learning avoding noises in synthetic triplets, we equip EntailE with an entity contrast module which mainly contributes to inductive CSKG completion. 
\end{itemize}
Experiments on CSKG completion tasks demonstrate the effectiveness of our method in both the transductive and inductive settings.
\section{Related Work}

\begin{table}[t]
    \centering
    \renewcommand{\arraystretch}{1.2}
    \caption{Definitions.}
    \label{tab:notation}
    \begin{tabular}{l|c}
    \toprule
    Terminology  & Definition \\
    \midrule
    Knowledge Triplet & (subject, relation, object)\\
    Named Entity & Real-world object, such as a person or location \\
    Predicate & Verbs, verbs with associated preposition \\
    Argument & Noun phrases, prepositional phrases, and adjectives\\
    Event & Composed by (\textit{predicates}, the set of \textit{arguments}) \\
    Hypernymy & A term is of supertype relation to another term \\
    Hyponymy & A term is of subtype relation to another term \\
    Textual Entailment & The truth of one text fragment follows another text \\
    \bottomrule
    \end{tabular}    
\end{table}

\subsection{CSKG Completion}
There are extensive research efforts in knowledge graph completion on factual knowledge graphs, like WordNet \cite{DBLP:journals/cacm/Miller95wordnet}, Freebase \cite{DBLP:conf/sigmod/BollackerEPST08}, YAGO \cite{Suchanek2007Yago}, and DBPedia \cite{DBLP:conf/semweb/AuerBKLCI07dbpedia}, in which the entity nodes are words or short phrases and graph structures are much dense. Unlike these knowledge graphs, commonsense knowledge graphs are with high-level semantic nodes and discourse relations, such as ConceptNet \cite{speer2017conceptnet}, ATOMIC \cite{sap2019atomic}, TransOMCS \cite{2020TransOMCS}, and ASER \cite{zhang2020aser}.


To address the sparsity problem, Malaviya et al. \cite{malaviya2020commonsense} and InductivE \cite {wang2020inductive} make use of synthetic edges and graph neural networks between nodes to enrich the entity representation learning. COMET \cite{bosselut2019comet} generates commonsense contextual nodes for completion with finetuned transformer over CSKG triplets. MICO \cite{su2022mico} converts the triplets into sentence pairs and adopts contrastive learning to improve the completion under inductive setting. Unlike them, we address the sparsity problem by utilizing the semantic plausibility with textual entailment relations. 


\subsection{Natural Language Inference}

Natural Language Inference (NLI) is a fundamental semantic task in natural language processing and involves reading a pair of sentences and judging the relationship between their meanings, such as entailment, neutral, and contraction. It is a task designed for evaluating the inference between two texts \cite{maccartney2008modeling}.

Learning sentence embedding for NLI has gained much attention in recent years. The representation benefits downstream tasks such as question answering \cite{wiebe2005annotating} and textual similarity \cite{marelli2014sick, cer2017semeval}. It captures high-level semantic information rather than only character or word embeddings from language models such as BERT \cite{sun2020self}. Recent methods for training sentence embeddings adopt finetuning transformers with NLI datasets \cite{bowman2015large, williams2018broad} under a supervised setting \cite{zhang2020semantics, sun2020self, wang2021entailment, gao2021simcse}. 

To better capture and represent the semantics of the nodes in CSKGs which are composed of entities, phrases, and events, we adopt SimCSE \cite{gao2021simcse} as a preprocessing tool to generate contextual embeddings in a unified form. The similarity score of the embeddings indicates the extent of the plausibility between the nodes. 

\begin{figure*}[t]
\centering
\includegraphics[scale=0.9,trim={0.0cm 0cm 0cm 0cm}]{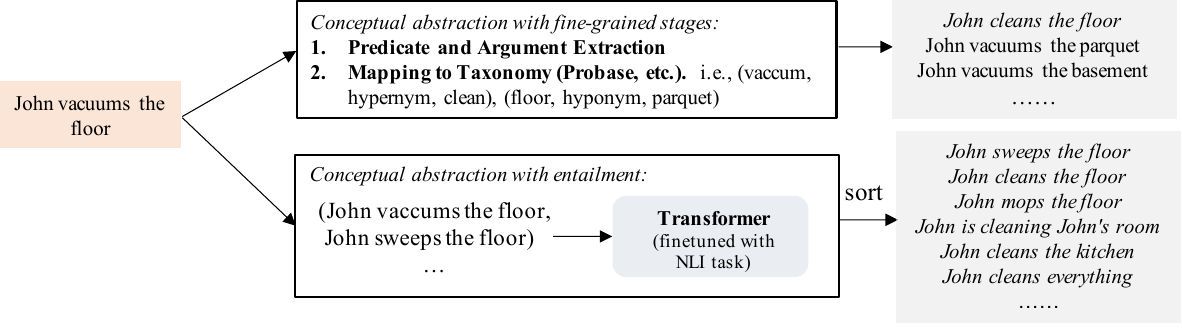}
\caption{Conceptual abstraction for finding plausibility consistent context node in ATOMIC. Texts in \textit{Italic} type are matched nodes in ATOMIC. }
\label{fig:plau}
\end{figure*}

\subsection{Plausibility of Commonsense Knowledge}
Semantic plausibility is an important component for commonsense related tasks, including commonsense reasoning \cite{bhagavatula2019abductive}, commonsense knowledge acquisition \cite{2020TransOMCS, yuenriching}, and commonsense question answering \cite{wang2023cat}. The plausibility of commonsense is generally across some appropriate of abstraction \cite{porada2021modeling}. Definitions for clarifying plausibility related items in introduced in Table \ref{tab:notation}.

Such conceptual abstraction can be realized by fine-grained stages: predicate and argument extraction, and mapping them to hypernymy or hyponymy from hierarchy lexical taxonomy such as WordNet or Probase \cite{van2009deriving, gong2016representing, he2020role}. Textual entailment is another line of work to model the plausibility, either in monotonic \cite{yanaka2019can, goodwin-etal-2020-probing, geiger-etal-2020-neural} or non-monotonic \cite{rudinger2020thinking} way. 

Recent method \cite{chen2022entailment} adopts textual entailment by inputting predicates to transformers finetuned over entailment recognition tasks, to alleviate the sparsity problem that predicate pairs may not co-occur in same corpus. Our method is similar to \cite{chen2022entailment}, we adopt a finetuned transformer to find entailment relations between node pairs, modeling the commonsense plausibility to densify the CSKG for better graph representation learning. \\

\section{Methodology}

\subsection{Task Definition}
Given a commonsense knowledge graph $\mathcal{G}:\{\mathcal{V}, \mathcal{R}\}$, $\mathcal{V}$ is the node set and $R$ is the relation set. It contains a group of knowledge triples $E=\{h,r,t\}, h,t \in \mathcal{V}, r \in \mathcal{R}$. CSKG completion task is to predict missing nodes in $(h,r,?)$ or $(?,r,t)$. The graph nodes are of free-form contexts.

\subsection{Entailment Relation for CSKG Nodes}
\label{method:entail}

Similar to \cite{chen2022entailment}, the conceptual abstraction with fine-grained stages on CSKG nodes also suffer from the data sparsity problem. After replacing the predicates or arguments with its abstracted or instantiated concepts, the new context nodes rarely exist in the current CSKG. This process brings in few useful triplets to densify the CSKG. \\

\noindent \textbf{Example 2:} Figure \ref{fig:plau} presents the example of conceptual abstraction from two kinds of methods. Fine-grained stages require extraction of predicate and argument structure and aligning them with concepts (e.g., hypernyms or hyponyms) in hierarchical lexical taxonomies such as WordNet \cite{DBLP:journals/cacm/Miller95wordnet} or Probase \cite{2012Probase}. Textual entailment is acquired by inputting node contexts into transformers finetuned on the NLI task, extracting embeddings and calculating similarity score. 

From the textual entailment method, the entire node contexts are viewed as input to the transformer finetuned over the NLI task. NLI task captures the entailment relation between two texts, which naturally facilitates the processing of node contexts in various forms. Specifically, we adopt a transformer $\mathbf{T}$\footnote{https://huggingface.co/princeton-nlp/sup-simcse-roberta-large} fine-tuned on NLI datasets from \cite{gao2021simcse} to extract the node embeddings. 

For every node $v \in \mathcal{V}$, the context $c$ of the node is the input to $\mathbf{T}$. The pooling representation of the last hidden layer is extracted as embedding representation for the node. Given node context pairs $\{c_i, c_j\}$, the similarity of them is defined as the cosine similarity between their representations:
\begin{equation}
    e_i = {\rm{Pool}}(\mathbf{T}(c_i)), e_j = {\rm{Pool}}(\mathbf{T}(c_j)),
\end{equation}
\begin{equation}
    sim(c_i, c_j) = \cos(e_i, e_j),
\end{equation}
The similarity score indicates the extent of the semantic plausibility between the two nodes. Each node in the graph gets an entailed node list after ranking the similarity score in descending order. 

\begin{figure*}[t]
\centering
\includegraphics[scale=0.85,trim={0.5cm 0cm 0cm 0cm}]{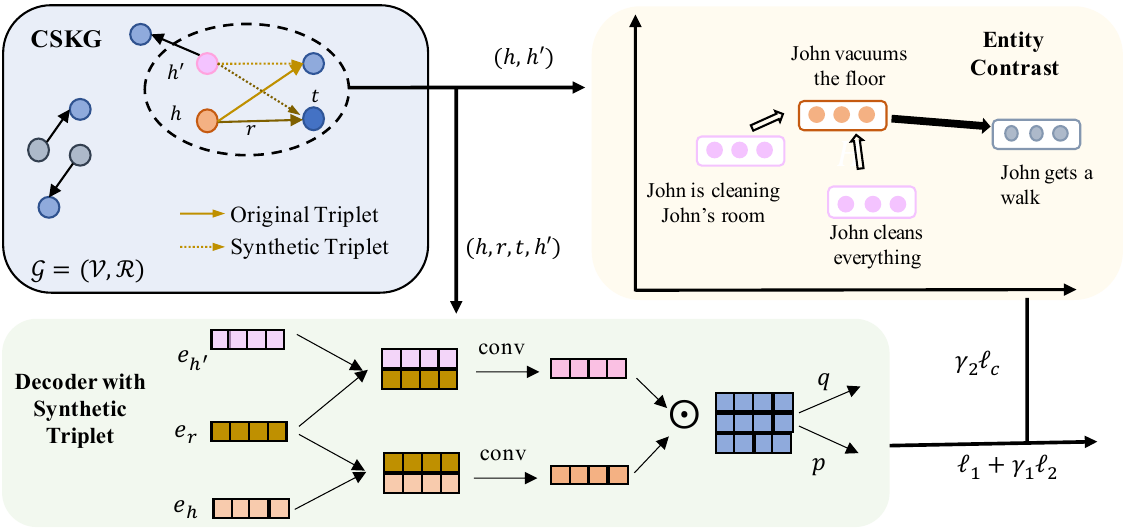}
\caption{The overview framework for EntailE. CSKGs are densified by the synthetic triplets. Entity contrast module groups the representation of source nodes with its entailed nodes. The triplet learning loss and entity contrast loss are trained alternatively.}
\label{fig:overview}
\end{figure*}

\subsection{EntailE for CSKG Completion}

The framework EntailE is composed of two parts: an decoder module and an entity contrast module. The overall framework is shown in Figure \ref{fig:overview}. \\

\noindent \textbf{Decoder}. We build based on the decoder model ConvTransE. The entity embedding matrix is of the CSKG node size. For entity embedding matrix initialization, we obtain the node embedding by inputting node contexts to a pre-trained language model (LM). The format of the input to the model is [CLS] + $c$ + [SEP], where $c$ is the natural language phrase represented by a node belonging to $\mathcal{V}$. The representation of the [CLS] token from the final layer of the pretrained LM as the node representation. Unlike Malaviya et al. \cite{malaviya2020commonsense}, we adopt an pre-trained LM BERT without fine-tuneing on CSKG triplets.

On top of the node embedding, we adopt ConvTransE \cite{shang2019end} as a decoder to score a knowledge triplet. The triplet embeddings $(e_h, e_r, e_t)$ are acquired with a look-up operator over entity embedding and relation embedding matrices. The ConvTransE decoder keeps the translational property between entity embedding and relation embedding by removing reshape operators compared to ConvE \cite{dettmers2018convolutional}. Assuming $K$ different kernels of size $2 \times W$ ($W$ as the width of a kernel), the output of $i$-th kernel at index $n$ of the output vector is given by:
\begin{equation}
    m_i(e_h, e_r)[n] = \sum_{\tau=0}^{W-1} w_i(\tau, 0)e_h(n + \tau) + w_i(\tau, 1)e_{r}(n + \tau),
\end{equation}
where $w_i$ is the trainable parameter in $i$-th kernel.
For scoring the triplet,
\begin{equation}
    p(e_h, e_r, e_t) = \sigma(f({\rm{vec}}(M(e_h, e_r))W_{c})e_t),
\end{equation}
where $M(e_h,e_r)$ is the matrix composed by outputs $\{m_1, ..., m_K\}$ from the kernels and $\rm{vec}$ operator converts the matrix into a vector. $W_{c}$ is a bilinear projection matrix and $f, \sigma$ are the non-linear functions. Finally, the module is optimized with a KvsALL training loss \cite{dettmers2018convolutional, wang2020inductive}:
\begin{equation}
    \ell_{1} = - \frac{1}{N} \sum_i(t_i \cdot \log(p_i) + (1-t_i) \cdot \log(1-p_i)),
\end{equation} 
where $N$ is the size of entity nodes and $i \in [1,N]$. $t$ is the true label and $p$ is the triplet score of size $N$. \\

\noindent \textbf{Synthetic Triplets}. We add synthetic triplets between original triplets and entailed nodes to alleviating the sparsity problem in the graph. 

Given an triplet $(h, r, t)$, for each head entity $h \in \mathcal{V}$, we get a set of top $k_1$ entailed node list $\{h^{'}_{1}, h^{'}_{2}, ..., h^{'}_{k_1}\}$ according to the calculation in section \ref{method:entail}. An synthetic triplet $(h', r, t)$ is constructed since $h$ entails $h'$ (an entailment edge between $h$ and $h'$). $h'$ is randomly sampled from the entailed node list $\{h^{'}_{1}, h^{'}_{2}, ..., h^{'}_{k_1}\}$.  Similar to the triplet score calculation of original triplets, for each synthetic triplet with $h'$, the score is:
\begin{equation}
    q(e_{h'}, e_r, e_t) = \sigma(f({\rm{vec}}(M(e_{h'}, e_r))W_{c})e_t),
\end{equation}
\begin{equation}
    \ell_{2} = - \frac{1}{N} \sum_i(y_i \cdot \log(q_i) + (1 - y_i) \cdot \log(1 - q_i)),
\end{equation} 

Note that to maintain the original CSKG structure, the tails $t$ in scoring triplets of $(h,r,t)$ remains unchanged with only synthetic triplets connected to $h'$. The training strategy aims to enhance the representation learning of node $h'$ with new triplet knowledge rather than transferring embedding from neighboring nodes. To balance the training of original triplets and synthetic triplets, a weighting parameter $\gamma_1$ is adopted to adjust $\ell_{2}$. \\

\noindent \textbf{Entity Contrast}. Since the synthetic triplets may induce noise as $(h',r,t)$ is not necessarily reasonable, we further design an entity contrast module for enhanced entity learning with entailment edge only. 
Similar to \cite{gao2021simcse}, we view entailed node pair $\{h, h'\}$ as positive pair and entailed node from other triplets as negative pairs. The entity contrast module groups the node embedding with semantic similarity while push away from irrelevant nodes.  For each node $h$, $h'$ is randomly sampled from its entailed node list $\{h^{'}_{1}, h^{'}_{2}, ..., h^{'}_{k_2}\}$. Node embedding is updated with a contrastive learning loss for $i$-th sample in a batch:

\begin{equation}
  \ell_{c} = -\log \frac{e^{sim(e_{h_i}, e_{h_{i}^{'}}) / \tau}}{\sum_{j=1}^{L} e^{sim(e_{h_i},  e_{h_{j}^{'}}) / \tau}}, 
\end{equation}
where $L$ is the batch size and $\tau$ is the temperature. $sim$ defines the similarity score, which can be calculated by dot product or cosine similarity. 


The final training loss is a combination of the three:
\begin{equation}
    \mathcal{L} = \ell_{1} + \gamma_1 \ell_{2} + \gamma_2 \ell_{c},
\end{equation}
where $\gamma_1$ and $\gamma_2$ are weight parameters of the loss. 

\section{Experiments}
This section introduces the evaluation metric for CSKG completion, benchmark datasets, baseline methods, and implementation details.

\subsection{Evaluation}
We evaluate the CSKG completion task with standard evaluation metrics for link prediction, including Hits@1, Hits@3, Hits@10, and Mean Reciprocal Rank (MRR).
\begin{equation}
    {\rm Hits}@n = \frac{1}{|Q|}\sum^{|Q|}_{i=1}\mathbb{I}(rank_i \leq n),
\end{equation}
\begin{equation}
    {\rm MRR} = \frac{1}{|Q|}\sum^{|Q|}_{i=1}\frac{1}{rank_i},
\end{equation}
The ${\rm Hits}@n$ metric evaluates the average rank of $Q$ queries. The indicator function $\mathbb{I}$ equals 1 if the rank is in top $n$ else 0. The MRR metric evaluates the average reciprocal ranking of all queries. 

For the evaluation setting, we apply both general setting and inductive setting \cite{wang2020inductive}. The general setting includes both transductive and inductive setting. \\
\noindent \textbf{Transductive Setting}. The CSKG completion is defined as predicting missing triplets $E'=\{(h,r,t)|(h,r,t) \notin E, h \in \mathcal{V}, t \in \mathcal{V}, r \in \mathcal{R}\}$. $E$ is the seen triplets in training split. $\mathcal{V}$ is the set of all seen nodes in the training split. \\
\noindent \textbf{Inductive Setting}. Inductive CSKG completion is defined as predicting missing triplets $E'' = \{(h,r,t)\}|(h,r,t) \notin E, h \in \mathcal{V}'$ or $t \in \mathcal{V}', r \in \mathcal{R}$, where $\mathcal{V}' \cap \mathcal{V} = \varnothing$ and $\mathcal{V}' \neq \varnothing$. 

During test stage, the score calculation for $(h,r,t)$ is $s = p(e_h, e_r, e_t)$. To see if entailed nodes $h'$ can benefit the head node $h$ in finding its tails given $r$, we denote a new score calculation as:
\begin{equation}
    s = {\rm{avg}} (p(e_h, e_r, e_t) + q(e_{h'}, e_r, e_t)),
\end{equation}
where ${\rm{avg}}$ is the average. For simplicity, we use the top one ranked entailed node. We denote the evaluation on test split with the new score calculation as EntailE$^{\ddagger}$

\subsection{Dataset}
Following \cite{wang2020inductive}, we conduct experiments on two standard benchmark datasets sampled from two typical CSKGs (i.e., ConceptNet \cite{speer2017conceptnet} and ATOMIC \cite{sap2019atomic}). Details of the distribution is in Table \ref{tab:dist}. Both of the CSKGs are sparse with average in-degree is 1.31 (ConceptNet) and 2.58 (ATOMIC) respectively. \\
\noindent \textbf{CN-82K}. CN-100K\footnote{https://ttic.uchicago.edu/\textasciitilde kgimpel/commonsense.html} is the version containing the Open Mind Common Sense (OMCS) entries from ConceptNet. CN-82K is a uniformly sampled version of the CN-100K dataset \cite{wang2020inductive}. Train/valid/test instances of the general setting are 81,920/10,240/10,240. Train/valid/test instances of the inductive setting are 81,920/9,795/9,796. \\
\noindent \textbf{ATOMIC}. The dataset split was created to make the set of seed entities manually exclusive between the training and evaluation splits from the ATOMIC \cite{malaviya2020commonsense}. In node context, subjects such as `PersonX' and `PersonY' are converted to `John' and `Tom'. Train/valid/test instances of the general setting are 610,530/87,707/87,709. Train/valid/test instances of the inductive setting are 610,530/24,358/24,488. \\

\begin{table}[t]
    \centering
    \caption{Distribution of CSKG.}
    \label{tab:dist}
    \begin{tabular}{l|p{1cm}p{1cm}p{1cm}p{1cm}p{1.2cm}}
    \toprule
    Dataset & Entities & Relations & Avg In-Degree & Avg Words & Unseen Entity(\%) \\
    \midrule
    CN-82K & 78,334 & 34 & 1.31 & 3.93 & 52.3 \\
    ATOMIC & 304,388 & 9 & 2.58 & 6.12 & 37.6 \\
    \bottomrule
    \end{tabular}    
\end{table}

\subsection{Baseline Methods}

The first line of baseline methods are applied in factual KG embedding learning. These methods stresses on modeling the multi-relational structure between head and tail nodes. These methods are classified into two categories, non-neural methods and neural methods. The non-neural methods use simple vector based operations for computing triplet score: \\
\noindent \textbf{DistMult} \cite{yang2015embedding}. DistMult models the three-way interactions between head entities, relation, and tail entities focusing on modeling symmetric relations. It adopts a basic bilinear scoring function based on dot product between the triplet embeddings: $<e_h, e_r, e_o>$. \\
\noindent \textbf{ComplEx} \cite{trouillon2016complex}. ComplexE extends DistMult by introducing complex embeddings so as to better model asymmetric relations. It scores the triplet in complex vector space with a Hermitian product: $Re(<{e_h, e_r, \bar{e}_o}>)$.  \\
\noindent \textbf{RotatE} \cite{sun2018rotate}. RotatE defines each relation as a rotation in the complex vector space with each $|e_r|=1$. The triplet scoring function is $-\|e_h \circ e_r - e_t\|^{2}$, where $\circ$ is a Hadmard product. \\

\begin{table*}[t]
    \renewcommand{\arraystretch}{1.2}
    \caption{Main results on CSKB completion under the general setting. }
    \label{Ret:1}
    \centering
    \begin{tabular}{p{2.3cm}|p{1.5cm}p{1.15cm}p{1.15cm}p{1.3cm}|p{1.15cm}p{1.15cm}p{1.15cm}p{1.15cm}}
    \toprule
    \multirow{2}{*}{Model}  & \multicolumn{4}{c|}{CN-82K} & \multicolumn{4}{c}{ATOMIC} \\
      & MRR & Hits@1 & Hits@3 & Hits@10 & MRR & Hits@1 & Hits@3 & Hits@10 \\
    \midrule
    DistMult & 2.80 & - & 2.90 & 5.60 & 12.39 & 9.24 & 15.18 & 18.30 \\
    ComplEx & 2.60 & - & 2.70 & 5.00 & 14.24 & 13.27 & 14.13 & 15.96 \\
    RotatE & 5.71 & - & 6.00 & 11.02 & 11.16 & - & 11.54 & 15.60 \\
    \midrule
    COMET &  - & - & - & - & 4.91 & 0.00 & 2.40 & 21.60 \\
    ConvE & 8.01 & - & 8.67 & 13.13 & 10.07 & 8.24 & 10.29 & 13.37 \\
    ConvTransE & 6.06$^{*}$ & 4.02$^{*}$ & 6.40$^{*}$ & 9.81$^{*}$ & 12.94 & 12.92 & 12.95 & 12.98 \\
    GCN+ConvTransE & 7.28$^{*}$ & 4.43$^{*}$ & 7.72$^{*}$ & 12.87$^{*}$ & 13.12 & 10.70 & 13.74 & 17.68\\
    ConvTransE$^{\dagger}$ & 12.87$^{*}$ & 8.41$^{*}$ & 13.88$^{*}$ & 21.63$^{*}$ & 12.33 & 10.21 & 12.78 & 16.20 \\
    \midrule
    MICO & 10.62 & 5.17 & 11.32 & 21.44 & 3.69 & 1.62 & 3.52 & 7.28 \\
    Malaviya et al & 16.26 & - & 17.95 & 27.51 & 13.88 & - &  14.44 & 18.38 \\
    \midrule
    EntailE($\gamma_2$=0.0) & 16.03$_{\uparrow 3.16}$ & 10.74$_{\uparrow 2.33}$ & 17.63$_{\uparrow 3.75}$ & 26.54$_{\uparrow 4.91}$ & \textbf{14.78}$_{\uparrow 2.45}$ & \textbf{11.60}$_{\uparrow 1.39}$ & \underline{15.47}$_{\uparrow 2.69}$ & \underline{20.66}$_{\uparrow 4.46}$ \\
    EntailE$^{\ddagger}$($\gamma_2$=0.0) & \underline{16.77}$_{\uparrow 3.90}$ & \textbf{11.30}$_{\uparrow 2.89}$ & \textbf{18.61}$_{\uparrow 4.73}$ & 27.69$_{\uparrow 5.06}$ & \underline{14.67}$_{\uparrow 2.34}$ & \underline{11.37}$_{\uparrow 1.16}$ & \textbf{15.58}$_{\uparrow 2.80}$ & \textbf{20.77}$_{\uparrow 4.57}$ \\
    \midrule
    EntailE($\gamma_2$=1.0) & 16.36$_{\uparrow 3.10}$ & 10.76$_{\uparrow 2.35}$ & 17.92$_{\uparrow 4.04}$ & \underline{27.72}$_{\uparrow 6.09}$ & 13.48$_{\uparrow 1.15}$ & 10.40$_{\uparrow 0.19}$ & 14.05$_{\uparrow 1.27}$ & 19.29$_{\uparrow 3.09}$ \\
    EntailE$^{\ddagger}$($\gamma_2$=1.0) & \textbf{16.92}$_{\uparrow 4.05}$ & \textbf{11.30}$_{\uparrow 2.89}$ & \underline{18.46}$_{\uparrow 4.58}$ & \textbf{28.16}$_{\uparrow 6.53}$ & 13.37$_{\uparrow 1.04}$ & 10.36$_{\uparrow 0.15}$ & 13.92$_{\uparrow 1.14}$ & 19.03$_{\uparrow 2.83}$ \\
    \bottomrule
    \end{tabular}
\end{table*}

\begin{table}[t]
    \renewcommand{\arraystretch}{1.2}
    \caption{Results on CSKG completion under the inductive setting.} 
    \label{Ret:2}
    \centering
    \begin{tabular}{p{1.95cm}|p{1.0cm}p{1.2cm}|p{0.8cm}p{0.9cm}}
    \toprule
    \multirow{2}{*}{Model} & \multicolumn{2}{c|}{CN-82K} & \multicolumn{2}{c}{ATOMIC} \\
      & MRR & Hits@10 & MRR & Hits@10 \\
    \midrule
    ConvE & 0.21 & 0.40 & 0.08 & 0.09 \\
    RotatE & 0.32 & 0.50 & 0.10 & 0.12 \\
    ConvTransE$^{\dagger}$ & 12.34 & 20.75 & 0.01 & 0.13 \\
    Malaviya et al. & 12.29 & 19.36 & 0.02 & 0.07 \\
    \midrule
    EntailE($\gamma_2$=0.0) & 14.92$_{\uparrow 2.63}$ & 25.14$_{\uparrow 4.39}$ & 1.03$_{\uparrow 1.02}$ & 2.11$_{\uparrow 1.98}$\\
    EntailE$^{\ddagger}$($\gamma_2$=0.0) & \underline{16.34}$_{\uparrow 4.00}$ & 26.83$_{\uparrow 6.08}$ & 1.47$_{\uparrow 1.46}$ & 2.88$_{\uparrow 2.81}$\\
    \midrule
    EntailE($\gamma_2$=1.0) & 15.98$_{\uparrow 3.64}$ & \textbf{26.97}$_{\uparrow 6.22}$ & \underline{1.71}$_{\uparrow 1.70}$ & \underline{3.85}$_{\uparrow 3.72}$ \\
    EntailE$^{\ddagger}$($\gamma_2$=1.0) & \textbf{16.58}$_{\uparrow 4.24}$ & \underline{26.93}$_{\uparrow 6.18}$ & \textbf{2.38}$_{\uparrow 2.37}$ & \textbf{5.12}$_{\uparrow 4.99}$ \\
    \bottomrule
    \end{tabular}
\end{table}

\noindent Neural methods include: \\
\noindent \textbf{ConvE} \cite{dettmers2018convolutional}. ConvE models the interactions between entities and relations with convolutional and fully-connected layers. The scoring function is defined by a convolution over 2D shaped embeddings: $\phi_r(e_h, e_t) = f({\rm vec}(f([\overline{e_h};\overline{e_r}] \ast \omega))\mathbf{W})e_t$, where $\overline{e_h}$ and $\overline{e_r}$ are a 2D reshapeing of $e_h$ and $e_r$, $\omega$ are the filters in a 2D convolutional layer. \\
\noindent \textbf{ConvTransE} \cite{shang2019end}. ConvTransE is designed as a decoder model based on ConvE with the translation property of TransE \cite{bordes2013translating}: $e_h + e_r \approx e_t$. The convolution filters are operated on $e_h$ and $e_r$ separately and then added. Aligning the output from all filters yield a matrix $\mathbf{M}(e_h, r_r)$. Final scoring function is $\phi(e_h, e_r) = f({\rm vec}(\mathbf{M}(e_h, e_r))W)e_t$, where $W$ is a matrix for linear transformation.\\

Another line of baseline methods considers the semantics in CSKG node contexts. Specifically, they are: \\
\noindent \textbf{Malaviya et al.} \cite{malaviya2020commonsense}. It initializes embedding with fine-tuned transformer. The embedding is concatenated with graph embedding extracted by GCN \cite{kipfsemi} for further learning. It adds synthetic edges over the entire graph with semantically similar meanings generated from the fine-tuned transformer to enhance graph embedding learning. 


\noindent \textbf{COMET} \cite{bosselut2019comet}. It is a generative method for training the language model on a set of knowledge triplets from CSKG. ATOMIC and ConceptNet are the CSKG sources. The training loss is to maximize the conditional loglikelihood of predicting tail node tokens in the knowledge triplet. 

\noindent \textbf{MICO} \cite{su2022mico}. It fine-tunes transformers to generate relation-aware node embedding, and scores the triplet with pairwise cosine similarity. The method takes in node and relation contexts as input, and  generating semantic representations for them. 

\subsection{Implementation Details}
We conduct experiments with BERT-large \cite{kenton2019bert} as the entity embedding initialization for the encoder-decoder module. Entity and relation embedding dimensions are set as 1024. For the convolutional decoder, we use 200 kernels of size $\{2 \times 5\}$ with ConvTransE. 

For extracting the entailed nodes, entity embedding is first extracted from the transformer fine-tuned with the NLI dataset. Then cosine similarity is calculated between each entity embedding with all other entities on the graph. For entailed edge, $\gamma_1$ is set as $\{0.3,0.5,1.0\}$ and $k_1$ is set as $\{1, 5,10,15\}$. For entity contrast, $\gamma_2$ is set as $\{0.0, 1.0, 2.0\}$ and $k_2$ is set as $\{5, 10, 15, 20\}$. The temperature of contrastive loss $\ell_{c}$ is set as 0.07. The training loss is optimized by Adam \cite{kingma2015adam} optimizer. For model training, the learning rate is set as 5e-5 on CN-82K and 3e-5 on ATOMIC. The batch size is set as 128.

We apply an alternatively optimized training schedule, in which one step we optimize the entire model with $\ell_{1}+\gamma_1\ell_{2}$, and next step we update the entity embedding matrix with $\gamma_2\ell_{c}$ only. For $\gamma_2=0$, the model is optimized with $\ell_{1}+\gamma_1\ell_{2}$ only. The average running time is 10 hours for each experiment on CN-82K and 72 hours on ATOMIC with one RTX A6000 GPU.

\section{Result Analysis}
In this section, we present the main results of CSKG completion under the general setting and inductive setting separately. Then we analyze the case study, as well as ablation study on  $k_1, \gamma_1, k_2, \gamma_2$. 

\subsection{Main Results}
The main results of CSKB completion under the general setting are shown in Table \ref{Ret:1}. In the table, the best results are in \textbf{bold-faced} and second-best results are \underline{underlined}. For CN-82K, $\gamma_1$=1.0, $k_1$=5, $k_2$=15. For ATOMIC, $\gamma_1$=0.5, $k_1$=5, $k_2$=10 in EntailE. EntailE$^{\ddagger}$ means averaging score from entailed nodes in evaluating with same model reported with EntailE. ConvTransE$^{\dagger}$ means entity embeddings are initialized with BERT-large. ConvTransE$^{\dagger}$ and EntailE has the same model size. $\uparrow$ indicates the performance gain of EntailE compared with ConvTransE$^{\dagger}$. Scores with $^{*}$ are from our implementation.

Previous vector-based methods developed on factual KG (DistMult, ComplEx, and RotatE) perform better over ATOMIC than CN-82K. This is because 
the average degree of ATOMIC is 2.58,
which is denser than CN-82K with an average degree of 1.31. Completion by generation method COMET \cite{bosselut2019comet} results in even sparser graphs due to free-from contexts and leads to low accuracy in Hits@1, Hits@3, and MRR.

The neural method ConvTransE (with randomly initialized node embeddings) achieves comparable or worse performances compared to vector-based method RotatE. While with initialization from BERT, ConvTransE$^{\dagger}$ has obvious performance gain on both benchmarks. The pre-trained language model BERT provides semantic initialization capturing the diversity of node contexts. ConvTransE$^{\dagger}$ outperforms GCN+ConvTransE, which adopts GCN to capture graph structure. This shows that semantic understanding of context nodes in CSKG is also an important aspect for CSKG completion besides graph structure.

Malaviya et al. adds synthetic edges to address the sparsity problem. The synthetic edges contribute to the node embedding update without involving in triplet scoring in the decoder. The method achieves decent performance gain under the general setting. Unlike them, the textual entailment in EntailE brings in new connections between entity nodes with high-level semantic understanding, covering not only semantic similarity but also conceptual abstraction. In EntailE, the synthetic edges are involved in decoder triplet scoring, with direct supervision on the entity and relation updating. Results show that EntailE can outperform Malaviya et al. on both benchmarks, demonstrating that bringing in textual entailment contributes to the CSKG completion.




\begin{figure}[t]
\centering
\includegraphics[scale=0.57,trim={0.5cm 0cm 0cm 0.5cm}]{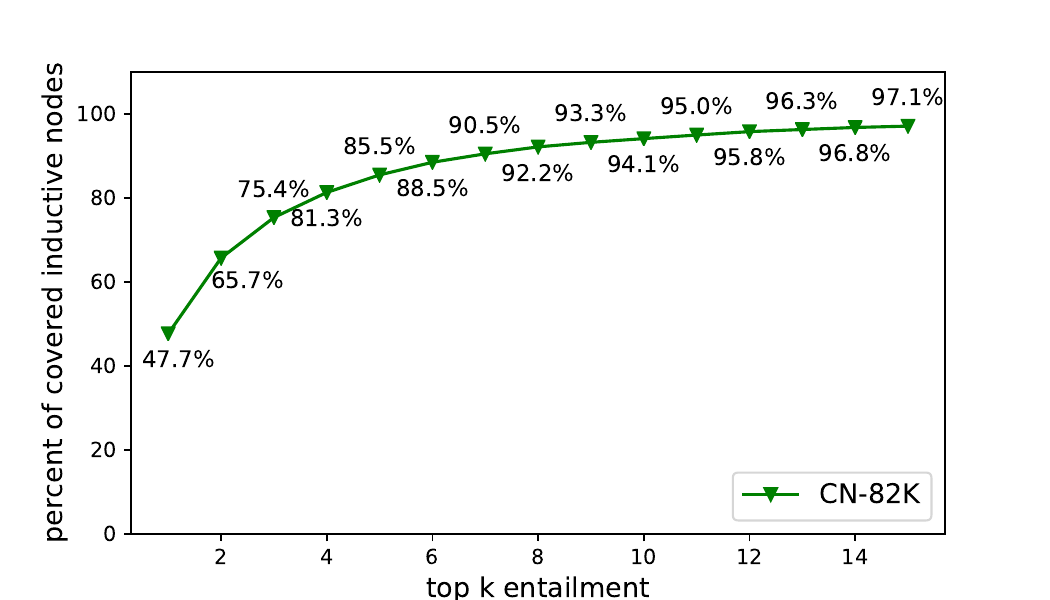}
\caption{Percentage of inductive CSKG nodes in the valid split covered by the entailed nodes of training split in CN-82K. $k$ is the number of entailed nodes for each node in the training split.}
\label{fig:inductive}
\end{figure}

\begin{figure}[t]
\centering
\includegraphics[scale=0.57,trim={0.5cm 0cm 0cm 0.5cm}]{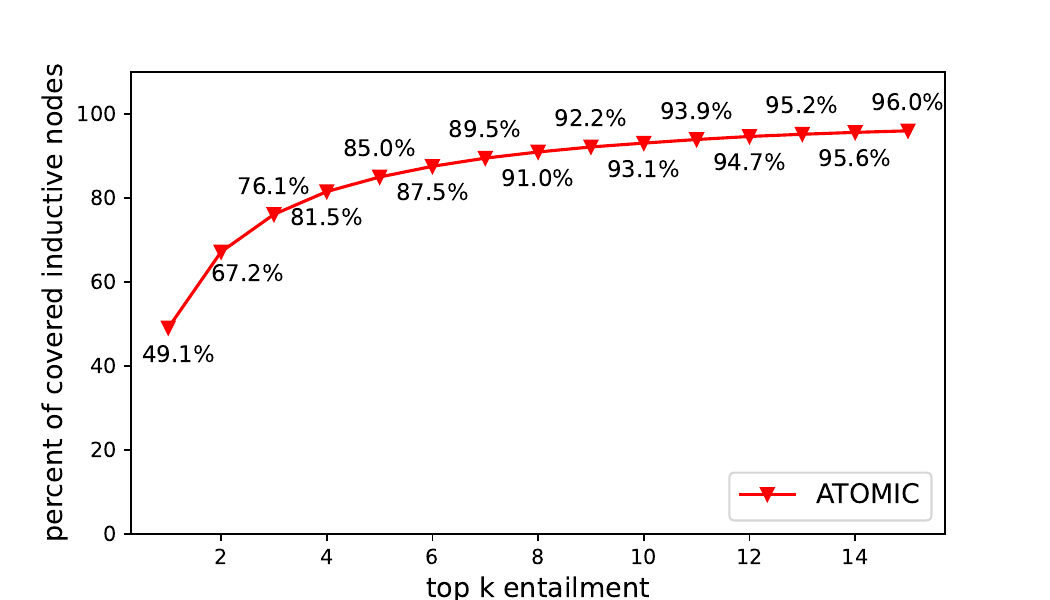}
\caption{Percentage of inductive ATOMIC nodes in the valid split covered by the entailed nodes of training split in ATOMIC. $k$ is the number of entailed nodes for each node in the training split.}
\label{append:1}
\end{figure}

Compared to a single encoder-decoder model without entailment relation (i.e., ConvTransE$^{\dagger}$)), our improvement is obvious in both benchmarks, with 5.50\% $\uparrow$ on Hits@10 over CN-82K and 4.57\% $\uparrow$ on Hits@10 over ATOMIC. Note that our model has the same parameter with ConvTransE$^{\dagger}$ but a different training strategy. CN-82K is more sparse than ATOMIC and thus adding synthetic edges based on textual entailment has more obvious benefits. It shows that mining plausibility from textual entailment are reasonable and can benefit CSKG completion significantly.

EntailE's entity contrast ($\gamma_2$=1.0) module has shown different effects on the two graphs, with an increase on CN-82K while a decrease on ATOMIC. The nodes in CN-82K are mostly short phrases while in ATOMIC are mostly long events. The textual entailment semantically finds more accurate conceptual abstraction in CN-82K.  
Entity contrast works for sparse and semantically simple CSKG but otherwise for relatively dense and semantically complex CSKG. When evaluated with entailment knowledge, EntailE$^{\ddagger}$ achieves performance gain on CN-82K while slight decrease on ATOMIC.

\begin{table}[t]
  \centering
  \renewcommand{\arraystretch}{1.2}
  \caption{Case study.}
  \label{tab:case}
  \begin{tabular}{lc}
  \toprule
  \multirow{2}{*}{Input} & Head $e_h$: John purchases the bike \\
  & Relation $e_r$: xWant \\
  \midrule
  Output & Predicted Tails $e_t$: \\
  \midrule
    & \textbf{to ride the bike}  \\     \cline{2-2}
    & \textbf{to take it home} \\ \cline{2-2}
   ConvTranse$^{\dagger}$ & to drive around town \\ \cline{2-2}
    & \textbf{to get exercise} \\ \cline{2-2}
    & to dirve home \\ 
  \midrule
  \multirow{2}{*}{} & \textbf{to ride the bike} \\ \cline{2-2}
    & \textbf{to take it home} \\ \cline{2-2}
  EntailE  & \textbf{to go to the store} \\ \cline{2-2}
    & \textbf{to relax} \\ \cline{2-2}
    & \textbf{to leave the store} \\
  \bottomrule
  \end{tabular}
\end{table}

\subsection{Inductive Setting}

The results under inductive setting are shown in Table \ref{Ret:2}. The best results are in \textbf{bold-faced} and second-best results are \underline{underlined}. The inductive setting measures the generalizability of node representation on unseen nodes. Since isolated nodes have no relation information to use, mining implicit connection to seen nodes from node contexts becomes intuitive for learning inductive node representation. 

The distribution of inductive nodes in the valid split covered by entailment relation on CN-82K and ATOMIC is presented in Figure \ref{fig:inductive} and Figure \ref{append:1} respectively. It shows that most of the inductive nodes (around 85\%) appear in the top 5 entailed nodes of training nodes. Therefore, textual entailment can help connect unseen nodes to training nodes by adding adding synthetic edges. 

Our method achieves significant performance gain on both benchmarks compared to ConvTransE$^{\dagger}$. It demonstrates that entailment relation can also improve the generalizability of inductive node representation. Our method outperforms Malaviya et al. by a large margin under the inductive setting. This demonstrates that exploring the conceptual abstraction in node contexts can benefit the completion especially when isolated unseen nodes are involved. 

Another finding is that entity contrast ($\gamma_2$=1.0) can also contribute to inductive learning on both benchmarks. Specifically, entity contrast can bring 6.22\% $\uparrow$ and 3.85\% $\uparrow$ on hit@10 over CN-82K and ATOMIC under inductive setting. This is quite different from the general setting. A potential reason is the proportion of transductive and inductive edges varies in the two benchmarks. Overall, the entity contrast module can utilize the entailment relation for better node representation.


\subsection{Case Study}
An case study of example predictions on ATOMIC test set using EntailE is presented in Table \ref{tab:case}. Top five predictions for example triplets generated from ConvTransE$^{\dagger}$ and EntailE in ATOMIC test split. ConvTransE$^{\dagger}$ and EntailE has same training parameters and initilization for node embeddings. The main difference is that EntailE is equipped with textual entailment guided training strategy.  Reasonable predicted tails are in \textbf{bold} type.

Comparing the top five predicted tails from ConvTransE$^{\dagger}$ and EntailE, we can find that EntailE can generate more reasonable tails. John can ``ride'' instead of  ``dirve'' the bike. Semantic differences between words in node contexts are very challenging to distinguish and EntailE can work towards this direction for a step. 

Tails in \textbf{bold} type are reasonable ones but not necessarily in the labeled test split. It is interesting that though they are reasonable but all of them are not labeled as ground tails given the head and relation. Therefore, the prediction accuracy is potentially underestimated due to event node plausibility, since top-ranked tails can be reasonable but not fully labeled. 

\begin{table}[t]
    \centering
    \renewcommand{\arraystretch}{1.2}
    \caption{Ablation study of $\gamma_1$ on CN-82K. $k_1$=10, $\gamma_2$=0.0.}
    \label{abla:2}
    \begin{tabular}{l|cccc}
    \toprule
    $\gamma_1$ & MRR & Hits@1 & Hits@3 & Hits@10 \\
    \midrule
    w/o entail & 12.87 & 8.41	& 13.88	 & 21.63 \\
    \midrule
    0.3 & 14.51 & 9.58 & 15.88 & 24.30 \\
    0.5 & 14.72 & 9.75 & 16.12 & 24.42  \\
    1.0 & \textbf{15.51} & \textbf{10.15} & \textbf{17.25} & \textbf{26.06} \\
    \bottomrule
    \end{tabular}    
\end{table}
\begin{table}[t]
    \centering
    \renewcommand{\arraystretch}{1.2}
    \caption{Ablation study of $k_1$ on CN-82K. $\gamma_1$=1.0, $\gamma_2$=0.0.}
    \label{abla:3}
    \begin{tabular}{l|cccc}
    \toprule
    $k_1$ & MRR & Hits@1 & Hits@3 & Hits@10 \\
    \midrule
    w/o entail & 12.87 & 8.41 & 13.88 & 21.63 \\
    \midrule
    1 & 14.57 & 9.71 & 16.00 & 24.03 \\
    5 & \textbf{16.03} & \textbf{10.74} & \textbf{17.62} & 26.54 \\
    10 & 15.51 & 10.15 & 17.25 & 26.06 \\
    15 & 15.90 & 10.54 & 17.56 & \textbf{26.62} \\
    \bottomrule
    \end{tabular}
\end{table}

\begin{table}[t]
    \centering
    \renewcommand{\arraystretch}{1.2}
    \caption{Ablation study of $\gamma_2$ on CN-82K. $k_2$=20.}
    \label{abla:4}
    \begin{tabular}{l|cccc}
    \toprule
    $\gamma_2$ & MRR & Hits@1 & Hits@3 & Hits@10 \\
    \midrule
    w/o contrast & 15.76 & 10.37 & 17.52 & 26.41 \\
    \midrule
    1.0 & 16.18 & 10.48 & 17.79 & \textbf{27.73} \\
    2.0 & \textbf{16.44} & \textbf{10.82} & \textbf{18.31} & 27.69 \\
    \bottomrule
    \end{tabular}
\end{table}

\begin{table}[t]
    \centering
    \renewcommand{\arraystretch}{1.2}
    \caption{Ablation study of $k_2$ on CN-82K. $\gamma_2$=1.0.}
    \label{abla:5}
    \begin{tabular}{l|cccc}
    \toprule
    $k_2$ & MRR & Hits@1 & Hits@3 & Hits@10 \\
    \midrule
    w/o contrast & 15.76 & 10.37 & 17.52 & 26.41 \\
    \midrule
    5 & 15.31 & 9.92 & 16.77 & 26.00  \\
    10 & 15.89 & 10.38 & 17.48 & 26.84 \\
    15 & \textbf{16.26} & \textbf{10.76} & \textbf{17.92} & 27.72 \\
    20 & 16.18 & 10.48 & 17.79 & \textbf{27.73} \\
    \bottomrule
    \end{tabular}
\end{table}

\subsection{Ablation Study}

\subsubsection{$\gamma_1$ and $k_1$}
Ablation study results of $\gamma_1$ and $k_1$ are shown in Table \ref{abla:2} and Table \ref{abla:3}.

From Table \ref{abla:2}, larger $\gamma_1$ means larger weight to the added triplet in training loss. The result shows that $\gamma_1=1.0$ achieves the best performance among the sampled values. In addition, the results with $\gamma_1> 0.0$ are better than the result without any entailment relation introduced. This shows the entailment relation is of high quality, and can help find reasonable triplets which benefits the CSKG completion task. 

With the number of entailment nodes increasing, the noise will also increase. $k_1$ determines the density of the new graph after entailment relation is introduced. In Table \ref{abla:3}, the performance stops increasing when $k_1$ becomes larger than 5. The learning becomes stable with more synthetic triplets involved. Overall, the performance gain is robust with a wide range of $\gamma_1$ and $k_1$ on both datasets.

\subsubsection{$\gamma_2$ and $k_2$}

Ablation study results of $\gamma_2$ and $k_2$ are shown in Table \ref{abla:4} and Table \ref{abla:5}. In the experiments, training loss for synthetic edges is the same with setting $\gamma_1$=0.5 and $k_1$=5. From Table \ref{abla:4}, the weight of entity contrast loss has a slight influence on the final results. Introducing entity contrast can consistently benefit the completion task on CN-82K. 

From Table \ref{abla:5}, we increase $k_2$ from 5 to 20 and can find that adding entity contrast can further benefit EntailE. Larger $k_2$ introduces larger coverage of isolated nodes becoming ``seen'' as presented in Figure \ref{fig:inductive}. Therefore, more isolated node embeddings will be updated with reasonable weak supervision signals in the entity contrast module. 

Overall, selection of the hyperparameters validates our methods for CSKG completion, by both adopting synthetic edges and entity contrast with textual entailment.



\section{Conclusion}
In this work, we propose to introduce textual entailment in CSKG completion. As a method to mine the plausibility of commonsense knowledge between entity nodes, the textual entailment knowledge extracted from NLI related tasks covers similarity as well as conceptual abstraction. Based on the textual entailment, we design synthetic edges to densify the CSKG. To enhance the entity representation learning, we design an entity contrast module to enlarge the difference the representation by utilizing entailed nodes as weak supervision signal. Experiment results demonstrate that adding entailed relations can benefit both the CSKG completion under the transductive and inductive settings.

%
\IEEEpeerreviewmaketitle

\bibliographystyle{IEEEtran}
\bibliography{bibtex/custum}
%




\end{document}